\documentclass{article}

\usepackage[final,nonatbib]{neurips_2019}

\usepackage[utf8]{inputenc} 
\usepackage[T1]{fontenc}    
\usepackage{hyperref}       
\usepackage{url}            
\usepackage{booktabs}       
\usepackage{amsfonts}       
\usepackage{nicefrac}       
\usepackage{microtype}      

\usepackage{graphicx} 
\usepackage{caption} 
\usepackage{keyval}
\usepackage[ruled,vlined]{algorithm2e}
\usepackage{amsmath}
\usepackage{amssymb}

\usepackage{amsthm}

\newcommand{\idest}{{\it i.e.}, }
\newcommand{\exempli}{{\it e.g.}, }

\usepackage{color}
\usepackage{cleveref}

\usepackage{titlesec}
\titleformat{\paragraph}
{\normalfont\normalsize\bfseries}{\theparagraph}{1em}{}
\titlespacing*{\paragraph}
{0pt}{3.25ex plus 1ex minus .2ex}{1.5ex plus .2ex}

\title{Explainable Semantic Mapping for First Responders}

\author{
Jean Oh, Martial Hebert, Hae-Gon Jeon, Xavier Perez, Chia Dai, Yeeho Song \\
    The Robotics Institute, Carnegie Mellon University \\
    Pittsburgh, PA, USA\\
  \texttt{\{hyaejino,mhebert,haegonj,xperez,chaid,yeehos\}@andrew.cmu.edu} \\
}

\begin{document}

\maketitle

\begin{abstract}
One of the key challenges in the semantic mapping problem in postdisaster environments is how to analyze a large amount of data efficiently with minimal supervision. To address this challenge, we propose a deep learning-based semantic mapping tool consisting of three main ideas. First, we develop a frugal semantic segmentation algorithm that uses only a small amount of labeled data. Next, we investigate on the problem of learning to detect a new class of object using just a few training examples. Finally, we develop an explainable cost map learning algorithm that can be quickly trained to generate traversability cost maps using only raw sensor data such as aerial-view imagery. This paper presents an overview of the proposed idea and the lessons learned. 
\end{abstract}

\begin{figure}[h!]
  \centering
  \begin{tabular}{cc}
  \includegraphics[width=0.44\columnwidth]{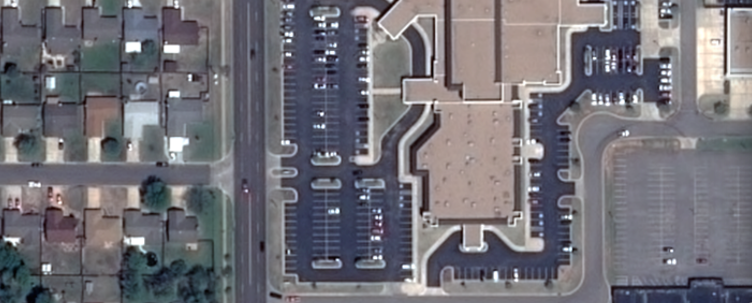} &
  \includegraphics[width=0.44\columnwidth]{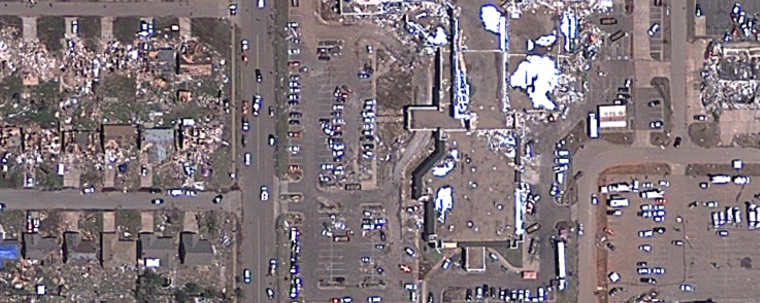}\\
  \textit{(a) Apr. 29, 2013 (Before)} &
  \textit{(b) May 22, 2013 (After)}\\
  \end{tabular}
  \caption{Aerial images \textit{(a)} before and \textit{(b)} after a tornado in Moore, Oklahoma, on May 20, 2013~\cite{googlecrisis2013}.}
  \label{fig:disaster}
\end{figure}

\section{Introduction}\label{sec:intro}
With the rise of affordable technologies such as low-cost cameras, drones and potentially ground robots with high terrainability, disaster relief is on the cusp of being transformed by new technological advances.  The use of drones following the earthquake in Ecuador in April 2016, for example, enabled building 2D or 3D images of the affected area by cross-stitching thousands of picture frames taken at various heights so that relief workers can visually assess damages. Based on these images, Raul Singh, a paramedic and founder of GlobalMedic~\cite{globalmedic2016} said, ``I [as a human] can tell you every building that is damaged on any street, assess the damage and decide the assets that are needed to deploy.'' As Kim Scriven, the manager of the Humanitarian Innovation Fund, noted in his BBC interview~\cite{BBC2014}, ``trying to harvest and filter the vast amounts of data generated by a disaster or conflict is the big nut that people are trying to crack, with the real challenge being to turn all of that data into information that humanitarian agencies can actually act on.'' Thus, the current challenge lies in semantic analysis of the large quantity of imagery data such as those shown in~\Cref{fig:disaster}. Such semantic analysis is mostly dependent on human effort, \exempli various humanitarian crowdsourcing~\cite{dhn,standby}. 

In this context, our work is motivated by the need for automatically building a semantic map from gigantic, raw imagery data to support first responders. To meet the level of urgency of disaster response, our proposed approach is specifically focused on semantic mapping approaches with unique desiderata: 1) that can be efficiently trained using a small number of labeled examples, 2) that scale well to handle big, high resolution images such as those aerial images collected by multiple drones from a large area affected by a disaster; 3) that can easily learn to classify new types of objects; and 4) that non-technical end users can easily use without special training. 

\section{Approach}

\begin{figure}[t!]
  \centering
  \includegraphics[width=0.7\textwidth]{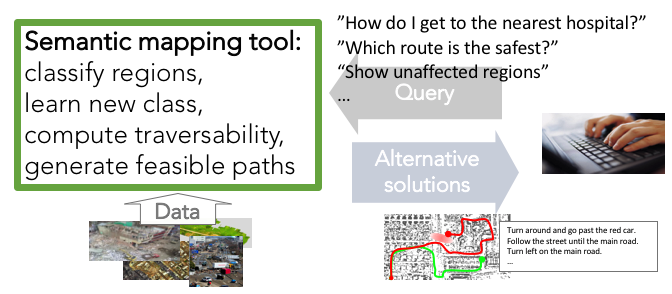}
  \caption{A scenario showing a use case of the proposed semantic mapping tool where the system assists a first responder by providing route options for dispatching and medical transportation operations in the affected areas.}
  \label{fig:scenario}
\end{figure}

We develop a semantic mapping system that can assist first responders as illustrated in~\Cref{fig:scenario}. The idea is to classify and analyze new semantic information about the affected areas from a large collection of aerial imagery data, and provide mapping and navigation services using the analyzed information. 
Our key objectives reflect the urgent needs of the first responders--namely, data-efficient training, adaptive learning, and intuitive interaction. With these objectives, the proposed approach includes three components described in the following subsections.

\subsection{Frugal semantic classification}\label{sec:frugal}

Generally speaking, when compared to image classification, pixel-wise classification requires not only a high-level understanding of the image to assign meaningful labels, but also precise localization of the label boundaries at a pixel level. This dual task is reflected in the network architecture, usually with a series of convolutions and pooling operations in the first layers until getting features with global information. The latter part of these architectures is then in charge of increasing the granularity of the output to a pixel level, by means of up-convolutions or interpolations. 

We study the following two groups of deep learning approaches that can potentially be trained fast while keeping the prediction accuracy level: Fully Convolutional Networks~\cite{long2014} and the network architectures based on the notion of hypercolumns~\cite{hariharan2014}.  

In PixelNet~\cite{bansal2017pixelnet}, the model builds on the hypercolumns idea~\cite{hariharan2014}, investigating the effect of using correlated information during training.  
They exploit the stochastic gradient descent assumption that training data are independent and identically distributed \textit{(i.i.d.)} sampled 
by randomly choosing a small subset of pixels per image during training. Additionally, instead of interpolating classifier scores, they interpolate the features in the deep layers that correspond to the subset of chosen pixels, and use a multi-layer perceptron in place of a linear classifier. They show that only using the 4\% of the pixels per image in every optimization step is comparable to using all the pixels in the image.

Sharing the intuition that image pixels are highly correlated, our approach builds on PixelNet~\cite{bansal2017pixelnet}. 
The fact that the PixelNet architecture can be trained with a fraction of labeled pixels in each image makes it suitable for scarcely labeled datasets; however, from the original work, we cannot infer the outcome of the training when parts of the ground truth are missing from the dataset. Every time that a picture is used in an optimization step, a new random selection of pixels is chosen. Hence, after the $80^{th}$ epoch described in the original work, most of the pixels in every image are likely to have been used during the training process. 

Our approach takes a step further and explores the performance of the PixelNet architecture with truncated data.
Our hypothesis is: since the pixels in near or similar regions are highly correlated, it is not necessary to have all the pixels labeled in the dataset to achieve competitive results. 
We make a simple but critical modification in the training process of PixelNet that allows the selection of which pixels are used (or not), based on arbitrary conditions.
Hence, as opposed to treating the unlabeled pixels from the scarce dataset as part of a meaningful class such as background or unlabeled, we take the design decision of not using them in the optimization process at all. 

\begin{figure*}[t!]
  \centering
  \includegraphics[width=\textwidth]{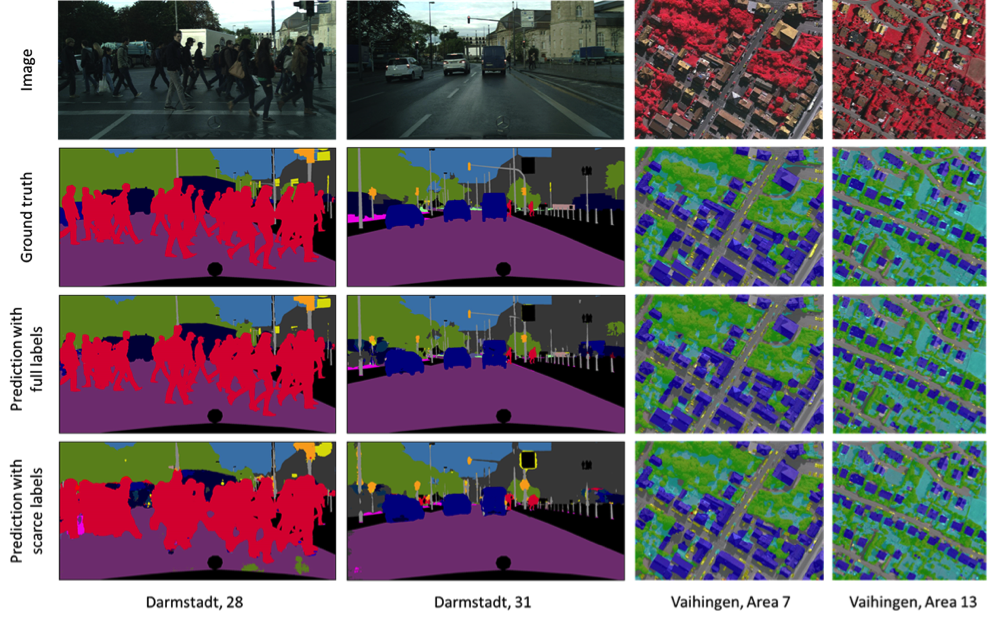}%
  \caption{Sample results on \textit{(left)} Cityscapes dataset~\cite{cityscapes} and \textit{(right)} WG II/4 Vaihingen dataset~\cite{isprs}.}\vspace{-3ex}%
  \label{fig:sample-results}%
\end{figure*}

To evaluate the approach, we use training datasets that are only partially labeled, \idest the majority of pixels does not have labels. The results on two datasets, aerial view and street view images, are promising. In the ISPRS Vaihingen dataset~\cite{isprs}, the model trained using only 35.29\% of labeled pixels performs comparably to that trained using full labels. Specifically, the accuracy drops only by 1.42\% when compared to the model trained using complete pixel-by-pixel labels. \Cref{fig:sample-results} shows qualitative results.

Through this work, we shed light on the algorithms using scarcely labeled data and report promising results. 
Additionally, we design our system as part of a semantic classification toolbox. This toolbox includes several common functionalities for pixel-level classification that can be shared for multiple network architectures. In addition to common tools such as data augmentation and preprocessing, it also provides pixel-level oriented functions such as the sampling of  different classes with different frequencies. We plan to share the toolbox with the research community.

\subsection{Learning a new class in unseen condition}\label{sec:new-class}

The goal of few-shot semantic segmentation is to learn a semantic map of the interested and unseen class given a few support images, their corresponding ground truth semantic map, and a query image to be predicted. Commonly, there are two types of data partitioning, one where training set contains the unseen classes but are labeled as background, and the other excluding all images containing unseen classes. The former suits the scenario where an annotator becomes interested in a sub-class of an existing class, \exempli green breaks into high vegetation and grass, or an annotator starts with a large background class and incrementally segments out the target classes of interest, \exempli adding a class of buildings out of background. In our experiments, a disaster scenario typically does not have pre-existing training samples; therefore, we choose the latter where a training set excludes all images containing test (query) classes.

\begin{figure}[ht!]
  \centering
  \includegraphics[width=0.8\columnwidth]{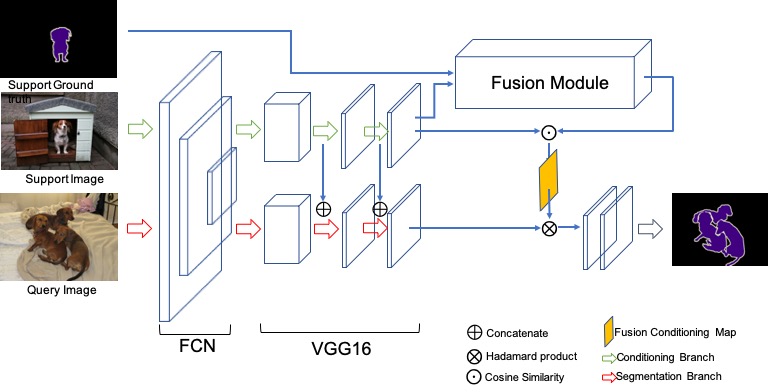}
  \caption{FuseNet: an overall architecture of one-shot example consisting of segmentation and conditioning branches~\cite{dai2019}}%
  \label{fig:FustNet}%
  \vspace{-10pt}%
\end{figure}%

\begin{figure}[b!]
  \vspace{-10pt}%
  \centering%
  \includegraphics[width=1\columnwidth]{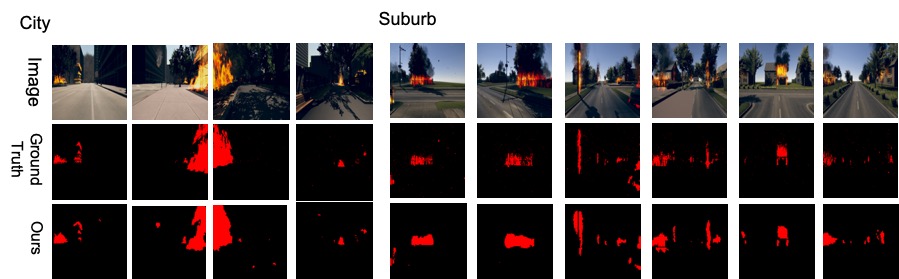}%
  \caption{A visualization of the FuseNet~\cite{dai2019} one-shot semantic segmentation on the city and suburban area fire dataset.}%
  \label{fig:kaist}%
\end{figure}%

Our work builds on the two-branch conditioning network and aims to tackle the  problems of overly-strict conditioning by extending one-shot to few-shot with more variety of support samples. In this paper, we share our high-level intuitions of our approach and refer the readers Our approach follows the idea from~\cite{dai2019} to which we refer the readers to for technical details. 

First, by combining the base segmentation network and the conditioning network into one common fully convolutional network~\cite{long2014}, we gain the benefit of the deep feature extraction and the reduced total number of parameters. Since the query and support samples come from a similar distribution within the same dataset, we hypothesize that it is sufficient to train one segmentation network for both branches. To ensure a similar distribution, we add an information flow from the conditioning branch to the segmentation branch by concatenating support-image features with query features in the last 2 layers of VGG16 as illustrated in~\Cref{fig:FustNet}.

Second, we create the Fusion Module, a plug-and-use module for masking intermediate features with a binary ground-truth map. Instead of masking out background from the input of support-image, leaving out other close-to-boundary information that may be helpful and creating an unnatural RGB image, we think that the local context around the new class can be used as a guidance. By fusing the mask with intermediate feature out of VGG, the network is able to look at the local contextual features of the new class.

Third, we extend from a one-shot learning to a few-shot learning by guiding with the sum of multiple conditioning maps weighted by a pairwise cosine similarity with the query feature map. One-shot often performs poorly when the support and query images have vastly different perspectives or background such as clutter. To improve the robustness and increase the mean Intersection over Union (IoU), we compute the fusion conditioning map for each support image. In contrast to~\cite{shaban2017one} and \cite{rakelly2018conditional} where they take the union of all K outputs of semantic maps,
or simply taking the average of fusion conditioning maps to use a centroid as a representation, we decouple the conditioning step from the query image. We, therefore, weigh each fusion map according to the distance, measured by cosine similarity, between their global feature map and the query global feature map. 

In summary, few-shot semantic segmentation aims to replace large amount of training data with only a few training samples, namely support samples. In this context, we extend one-shot learning to few-shot and propose a new network, FuseNet, that merges multiple feature vectors from support images and leverages cosine similarity as guidance to predict segmentation mask. We also explore the effects of number of support images quantitatively on Intersection over Union (IoU) and qualitatively on visualization. 
\Cref{fig:kaist} shows qualitative results demonstrating that the proposed approach was able to learn a new class, fire, using only a single training example.  

\subsection{Explainable semantic mapping and navigation}\label{sec:semantic-nav}

The idea of imitation learning has been successfully used for offroad navigation where semantically classified aerial-view images are used to generate traversability cost maps. Following the idea of using aerial-view images for maps, we develop a deep learning framework for generating navigation cost maps that can be customized for task-specific preferences. 
Leveraging deep convolutional neural networks, we aim to capture useful features that are beyond a set of predefined semantic labels. At the same time, we utilize the set of known labels to generate human-interpretable explanation for the paths generated by the proposed approach.

Here, we use a deep inverse reinforcement learning (IRL) algorithm~\cite{song2019} for learning various navigation cost maps using high-resolution, large-sized satellite/aerial images. We refer the readers to the original reference~\cite{song2019} for technical details of the deep IRL approach. Here, we briefly introduce the explainability of our approach. 

\begin{figure}
    \centering
    \includegraphics[width=0.98\textwidth]{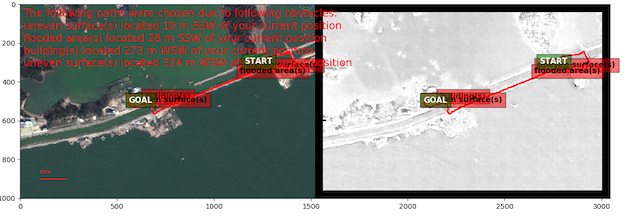}
    \caption{Given a user query ``Find a safe route to transport patient 1 to the nearest hospital,'' our system generates a safe traversability map (right) to find a path and explains why this path has been chosen.}%
    \label{fig:explain-route}%
    \vspace{-0pt}
\end{figure}%

For example,~\Cref{fig:explain-route} shows how semantic navigation using deep IRL can be used in the proposed semantic mapping services. In this example, our approach generates a feasible path given an aerial image and an end-user's navigation query ``find a safe route to transport patient 1 to the nearest hospital.'' The system explains that the path was chosen over a possibly shorter route due to the flooded area in the shortest path. We also note that, in this example, the label ``flooded area'' was learned using FuseNet, the few-shot learning algorithm.

\section*{Acknowledgement}
This work is in part supported by the Air Force Office of Scientific Research under award number FA2386-17-1-4660. This work should not be interpreted as representing the official policies, either expressed or implied, of the Army Research Laboratory of the U.S. Government. The U.S. Government is authorized to reproduce and distribute reprints for Government purposes notwithstanding any copyright notation herein.

\section{Conclusion}
Navigation in a postdisaster environment presents unique challenges due to drastic physical changes made in the affected environments. For instance, prior maps may no longer be valid, and normal definitions of traversability may not even be applicable to find safe paths. In this context, We discuss technical challenges in generating semantic navigation maps in support of first responders, focusing on how we can process a large amount of unlabeled aerial or satellite images. We present three core ideas of our proposed approach to tackle such challenges and share the findings from our work. 

\small

\bibliographystyle{plain} 
\bibliography{main} 

\end{document}